\newif\ifsinglecolumn  
\singlecolumnfalse
\newif\ifrelaxedv  
\relaxedvtrue
\newif\ifextendedv  
\extendedvtrue
\newif\ifbios 
\biosfalse

\ifsinglecolumn
\documentclass[journal,10pt,onecolumn,draftclsnofoot]{IEEEtran}   
\else
\documentclass[10pt,twocolumn,twoside]{IEEEtran}  
\fi

\newcommand{\jvspace}[1]{\ifrelaxedv\else\vspace{#1}\fi}

\usepackage{url}
\usepackage{graphicx}
\usepackage{mathtools}

\usepackage{amsmath}
\usepackage{amssymb}
\usepackage{color}
\usepackage{psfrag}
\usepackage{graphicx}
\usepackage{boxedminipage,wrapfig}

\usepackage[]{algorithm2e}
\usepackage{multirow}

\usepackage{lipsum}                     
\usepackage{xargs}                      
\usepackage[pdftex,dvipsnames]{xcolor}  
\usepackage[colorinlistoftodos,prependcaption]{todonotes}

\newtheorem{problem}{Problem}
\newtheorem{assumption}{Assumption}

\newcommand\ndots{\hbox to 1em{.\hss.\hss.}}
\graphicspath{{figures/},{}}

\newif\ifmargincomments 
\margincommentsfalse

\ifmargincomments

\newcommandx{\federico}[2][1=]{\todo[inline,linecolor=magenta,backgroundcolor=magenta!25,bordercolor=magenta,#1]{#2}}
\newcommandx{\tiago}[2][1=]{\todo[inline,linecolor=yellow,backgroundcolor=yellow!25,bordercolor=yellow,#1]{#2}}
\newcommand{\rev}[1]{#1}
\newcommand{\revtwo}[1]{{\color{blue}#1}}
\else

\newcommandx{\federico}[2][1=]{}
\newcommandx{\tiago}[2][1=]{}
\newcommand{\rev}[1]{{#1}}
\newcommand{\revtwo}[1]{#1}
\fi

\newcommand{\size}[1]{\ensuremath{s\left({#1}\right)}}
\newcommand{\data}[1]{\ensuremath{d({#1})}}
\newcommand{\ctime}[2]{\ensuremath{\tau_{#1}\left({#2}\right)}}
\newcommand{\cenergy}[2]{\ensuremath{C_{#1}^e\left({#2}\right)}}
\newcommand{\agent}[1]{\ensuremath{A_{#1}}}
\newcommand{\rate}[3]{\ensuremath{r_{{#1}{#2}}({#3})}}
\newcommand{\sol}{\ensuremath{\mathbb{S}}}
\newcommand{\tasks}{\ensuremath{\mathbb{T}}}
\newcommand{\required}{\ensuremath{\mathbb{R}}}
\newcommand{\dtime}{k}
\newcommand{\interferenceset}{I}
\newcommand{\interferencesets}{\mathbb{I}}

\begin{document}

\title {Multi-Robot On-site Shared Analytics Information and Computing}

 \author{Joshua Vander Hook\IEEEauthorrefmark{1}\thanks{\IEEEauthorrefmark{1} Corresponding author}, Federico Rossi, Tiago Vaquero, Martina Troesch, Marc Sanchez Net, Joshua Schoolcraft, Jean-Pierre de la Croix, and Steve Chien
 \thanks{All authors are with the Jet Propulsion Laboratory, California Institute of Technology, 4800 Oak Grove Dr., Pasadena (CA) 91101. Email: \{hook, tiago.stegun.vaquero, federico.rossi, martina.i.troesch, marc.sanchez.net, joshua.schoolcraft, jean-pierre.de.la.croix, steve.a.chien\}@jpl.nasa.gov}}

\maketitle
\begin{abstract} 

Computation load-sharing across a network of heterogeneous robots is a promising approach to increase robots capabilities and efficiency as a team in extreme environments. 
However, in such environments, communication links may be intermittent and connections to the cloud or internet may be nonexistent.
In this paper we introduce a communication-aware, computation task scheduling problem for multi-robot systems and propose 
an integer linear program (ILP) that  optimizes the allocation  of  computational  tasks  across a network of  heterogeneous  robots, accounting  for  the  networked robots’  computational  capabilities  and for  available  (and  possibly  time-varying)  communication links. 
We consider scheduling of a set of inter-dependent required and optional tasks modeled by a dependency graph.  We present a consensus-backed scheduling architecture for shared-world, distributed systems.
We validate the ILP formulation and the distributed implementation in different computation platforms and in simulated scenarios with a bias towards lunar or planetary exploration scenarios.  
\revtwo{Our results show that the proposed implementation can optimize schedules to allow a threefold increase the amount of rewarding tasks performed (e.g., science measurements) compared to an analogous system with no computational load-sharing. }

\end{abstract}

\section{Introduction}
\label{sec:intro}

Multi-agent systems hold great promise for science exploration in extreme
environments.  Correspondingly, there has been a proliferation of national
programs aimed at expanding multi-agent networked systems for caves
\cite{chung2019darpa}, oceans \cite{waterston2019ocean} and low earth
orbit \cite{kramer2008overview,pekkanen2019governing}.  These environments can
be considered the ``extreme edge'', far from the robust computation and
omnipresent communication networks of connected cities.  

In planetary exploration there is an emerging push toward more extreme environments,
and therefore multi-agent systems because single, flagship robots are limited to less-hostile
operating areas. Therefore, access to Recurring Slope Lineae or planetary caves
\ifextendedv
\cite{boston-frederick-et-al-2003,leveille2010lava,mcewen2014recurring}
\else
\cite{leveille2010lava,mcewen2014recurring}
\fi
may be possible with
multiple small, potentially expendable rovers.  Not surprisingly, we see potential 
systems being demonstrated in the Mars helicopter\cite{balaram2018mars}, and the
“PUFFER” rover (Pop-Up Flat-Folding Explorer Robots)
\ifextendedv
\cite{karras2017pop,davydychev2019design}. 
\else
\cite{davydychev2019design}. 
\fi
There is also evidence that
next-generation spaceflight computing employed on these systems will be more
like our current mobile devices \revtwo{
\cite{balaram2018mars,doyle2013hpc,powell2011enabling,mounce2016hpc,schmidt2017spacecubex}.} Finally, it is likely that, in the future, multiple collaborating robots, astronauts, and base stations will \emph{themselves} be a complex and time varying processing and communication network \revtwo{(see e.g. \cite{turchi2021system})}.

\begin{figure}[htb]
  \centering{
  \includegraphics[width=\linewidth]{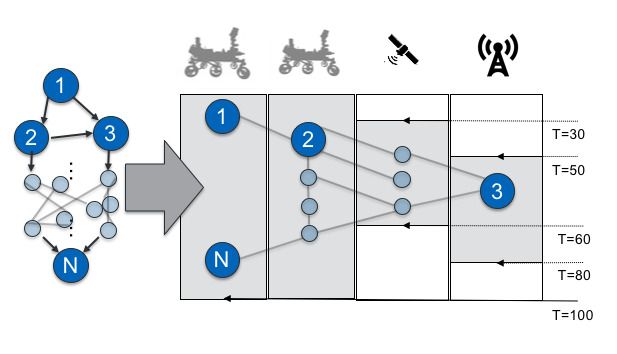} 
  {\jvspace{-2em}
   \caption{
    Illustrative MOSAIC scenario.
    A set of processing tasks (On the left as dependency graph) must
    be mapped to multiple assets with heterogeneous computing, communication,
    and energy capacities. Each asset is also available over a fixed
    time window due to terrain effects or orbital parameters.  
    The goal is to compute all the required tasks as quickly as
    possible.
    \label{fig:rovers}
  }
  }
}
\end{figure}

The key metric of these system concepts is throughput of observations and data. 
A compelling paradigm to increase the throughput of heterogeneous multi-robot
systems is \emph{computational load-sharing}: by allowing robotic agents to
offload computational tasks to each other or to a ``computational server''
(e.g., an overhead orbiter, a flagship rover, or a stationary lander),
computational load-sharing can give access to advanced analysis capabilities
to small, low-power rovers with limited on-board computing capabilities, or allow agents to do more \revtwo{memory- or cpu-intensive} work by leveraging nearby idle nodes. 
Previous work \cite{hook2018ICAPSws,hook2019ICAPS} has shown that computation
sharing in robotic systems with heterogeneous computing capabilities (e.g., Mars
exploration scenarios) can lead to significantly increases in system-level
performance and science returns. 

\revtwo{These self-reliant, edge robotic systems share commonalities that motivate our
study.  The first is an emphasis on energy conservation due to their remote,
self-sustaining design. The second is the possible use of heterogeneous
systems, in which some nodes contain more resources (power, computing,
communication, mobility, sensing, etc) than others.  The final factor is
intermittent and periodic loss of connectivity between nodes.  While the
benefit of edge computation supporting mobile phone networks continues to be well
investigated (see the highly influential \cite{cuervo2010maui}), the intermittent loss of
communications and time-varying position of the agents makes it more challenging to employ
these concepts directly. This is true because it is challenging to route
through a changing network, but also because  the source and destination changes over time
because agents are collaborating and assisting each other.  (Figure
\ref{fig:rovers}).  The resulting solution must tolerate partitions to the
network or long delays before data can be sent between nodes or back to a data
center.}

In this paper, we formalize the \textit{communication-aware 
computation task scheduling problem} and present an Integer Linear Program that
optimizes the allocation of computation and communication tasks to heterogeneous agents,
accounting for the computational capabilities and 
time-varying communication links.  Because data and computation are
shared among many devices, we dub the resulting local computation-sharing
network a MOSAIC (Multi-robot On-site Shared Analytics Information and Computing) network. 

We model and test with networks that use Delay- and Disruption-Tolerant
Networking (DTN) which provides transparent store-and-forward, multi-hop data
routing between arbitrary endpoints and negotiates intermittent interruptions
and delays in connectivity \ifextendedv
\cite{wyatt2017,cerf2007delay,burleigh2003delay}
\else
\cite{wyatt2017,burleigh2003delay}
\fi.
\revtwo{Unlike mobile phone networks which respond to consumer demand, in cooperative multi-agent networks,  
agents can explicitly share their goals and constraints with each other. }
Thus, we consider
the robots' intended actions as part of the scheduling problem so that the
robots can schedule data-intensive tasks when assistance is available.  
Our evaluation scenarios are biased towards multi-rover systems for Mars or the
Moon.  However, the results generalize to arbitrary time-varying
communication graphs, such as vehicles on known street routes or constellations
in orbit.  In a planetary exploration scenario, we show that distributed
computation can increase the amount of science performed \emph{threefold}
compared to an analogous system with no computational load-sharing. We show
that the solution includes intuitive results such as designated relay nodes and
``assembly line'' behaviors.


\subsection{Related Work} \label{sec:related_work}

The core computational problem addressed in this work is communication-aware task scheduling.
Task scheduling is known to be NP-complete
\ifextendedv
\cite{garey1979computers,ullman1975np};
\else
\cite{garey1979computers};
\fi
furthermore, while polynomial-time approximation schemes for the problem exist, to the best
of the authors' knowledge no such schemes are known for the task scheduling
problem when computing nodes have \emph{heterogeneous computational
capabilities}, i.e. the same task requires different computation runtimes on
different nodes
\ifextendedv
 \cite{Graham1979Optimization,Kwok1999Static}.
\else
 \cite{Graham1979Optimization}.
\fi
A large number of heuristic algorithms have been proposed to solve
the task scheduling problem. Heuristics may be classified as \emph{list
scheduling} heuristics (e.g., \cite{Sih1993DynamicLevelScheduling}), which rely
on greedily allocating tasks according to a heuristic priority task assignment;
\emph{clustering} heuristics (e.g., \cite{Yang1994clustering}), which identify
groups of tasks that should be scheduled on the same computing node; and
\emph{task duplication} heuristics, which duplicate some tasks to reduce
communication overhead (e.g., \cite{ahmad1994duplication}). In addition, a
number of \emph{guided random search} algorithms are available, including genetic algorithms
\cite{yu2006scheduling} and ant colony optimization algorithms
\cite{Chen09Scheduling}. See the survey in
\cite{Kwok1999Static} and introduction in \cite{Topcuoglu2002HEFT} for a
thorough review.

In particular, the heterogeneous earliest-finish-time (HEFT) heuristic
algorithm \cite{Topcuoglu2002HEFT} provides excellent performance for
heterogeneous task scheduling problems, and a number of variations of HEFT have
been proposed \cite{Tang2009HEFT,Canon2010RobHEFT,Tang2011SHEFT}.  However, the
HEFT algorithm and its derivatives generally assume that computation nodes are
able to perform all-to-all communication and that the availability of
communication links does not change with time; they also  do not capture access
contention or bandwidth constraints on communication links, and do not
accommodate \emph{optional} tasks which are not required to be scheduled but
result in a reward when added to the schedule.

Heuristic approaches are also used in model-based schedulers/temporal-planners that rely on activity-centric representations such as timeline-based modeling languages  \cite{chien-et-al-SpaceOps-2012} and the Planning Domain Definition Language (PDDL) \cite{pddl21,pddl22,pddl30}. Research on PDDL temporal planners for example has focused on domain-independent heuristics and has deployed planning systems to several robotics applications, especially those that require both planning and scheduling capabilities \cite{cashmoreetal2014,colesetal2019}.
One of the main state-of-the-art temporal planner is OPTIC \cite{benton2012temporal}; OPTIC not only reasons about actions' preconditions and effects to determine the set of actions required to achieve a given goal state, but also considers an action's temporal and resource constraints as well as soft state constraints (preferences) and continuous objective functions. Due to its generality in the input representation, we compare the performance of our approach with the OPTIC planner in Section \ref{sec:experiment:benchmark}.

Several heuristics are also available for the \emph{online} scheduling problem,
where computational tasks  appear according to a stochastic process, and are
not revealed to the scheduler in advance
\cite{tassiulas1992stability,dai2005maxpressure,TerekhovEtal2014JAIR}; recent work extends such
schedulers to accommodate communication latency constraints
\cite{Yang2018Scheduling}. However, online approaches generally perform poorly
compared to offline algorithms when the list of tasks to be executed is known
in advance or in batch, a typical scenario for robotic exploration missions;
in addition, even state-of-the-art online algorithms assume that
\textit{all-to-all communication between the computation nodes is available}.  In
contrast, the approach proposed in this paper 
does adapt to realistic time-varying communication constraints, explicitly represents
multi-hop communications between nodes, and accommodates optional
tasks, while offering sufficiently fast computation times to make the approach amenable
for field use, as we show.

\revtwo{
The problem of resource-aware scheduling in space application has seen a significant amount of interest in the adaptive space systems community. However, existing solutions tend to focus on reconfigurability \emph{within} an individual vehicle (see e.g. \cite{fayyaz2012adaptive}); solutions applicable to multi-agent systems generally assume that all-to-all communication is available \cite{liao2019caching}.
}

\subsection{Contribution}
Our contribution is threefold.
First, we design a task scheduling and task allocation algorithm based on integer programming that accounts for time-varying, bandwidth-constrained, multi-hop communication links and optional tasks, and that returns high-quality solutions quickly. We also provide a distributed implementation of the algorithm based on a shared-world, consensus-backed model.
Second, we validate the performance of the algorithm with extensive benchmarking on several hardware architectures, including embedded architectures such as PPC 750 and Qualcomm Flight, and with human-in-the-loop field tests.
Third, we explore and highlight emergent load-sharing behaviors produced by the scheduling algorithm, and we quantitatively show that sharing of computational tasks can result in significant increases in science throughput for a notional multi-robot mission.
Finally, we provide an open-source implementation of the core results for the community's use. 

Collectively, the results in this paper show that sharing of computational
tasks among heterogeneous agents is greatly enhancing of 
heterogeneous multi-agent architectures, resulting in higher utilization of
computational resources, lower energy use, and increased scientific throughput for a given
hardware architecture.

A preliminary version of this paper was presented at the 2019 International Conference on Automated Planning and Scheduling (ICAPS) conference \cite{hook2019ICAPS}.
In this extended version, we
(i) provide an in-depth discussion of the ILP problem and several additional extensions (including additional cost functions and first-order modeling of network interference),
(ii) rigorously show that a flooding-based algorithm can be used to provide a distributed
implementation of the scheduling algorithm for systems with moderate numbers of agents,
(iii) report extensive benchmarking results showing that the ILP can be solved
effectively on embedded hardware architectures suitable for robotic systems,
and (iv) present an extended discussion of experimental results.

\subsection{Organization}
The rest of this paper is organized as follows.
In Section \ref{sec:modelling}, we rigorously describe the multi-robot, communication-aware  computation task scheduling problem solved in the paper.
In Section \ref{sec:scheduler}, we provide a detailed description of the proposed scheduling algorithm.
In Section \ref{sec:experiment}, we present experimental results from a field test performed at Jet Propulsion Laboratory (JPL) and highlight a number of interesting emerging organization behaviors. We also report benchmarks showing that the proposed scheduling algorithm performs well on several embedded hardware architectures.
Finally, in Section \ref{sec:disc}, we draw conclusions and lay out directions for future work. 

\section{Problem Description}\label{sec:modelling}

\begin{figure}
  \centering
  \includegraphics[width=.95\columnwidth,trim={0.2cm 0 0 0}, clip]{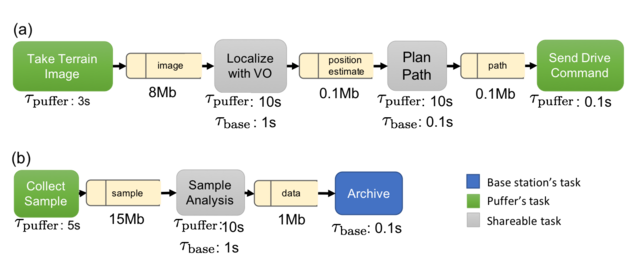}
  \caption{Notional software network for the PUFFERs rovers.}
  \label{fig:scenario_task_network}
\end{figure}

We now describe the communication-aware  computation task scheduling problem for heterogeneous multi-robot systems.


\paragraph{Tasks and Software Network}
\revtwo{ We wish to schedule a set \tasks~ of tasks, with $|\tasks|=M$ (that is, the number of tasks in \tasks~ is $M$).}
Computational tasks of interest can include, e.g. localizing a robot, computing a motion plan for a robot,
classifying and labeling the content of an image \cite{ono2016data,Higa2019VeeGer}, or estimating the spatial distribution of a phenomenon
based on point measurements from multiple assets \cite{tokekar2016sampletspn,kriging}.

Tasks may be \emph{required} or \emph{optional}. Required tasks,
denoted as $\required\subseteq\tasks$, must be included in the schedule.
\revtwo{Optional tasks $\tasks \setminus \required$ are each assigned a \emph{reward} score, denoted as $r(T)$ for each optional task $T\in\tasks\setminus\required$, which captures the value of including the task in a schedule.}

The output of each task is a \emph{data product}. Data products for task $T$ are denoted as \data{T}.
The size (in bits) of the data products are known a-priori as \size{T} for task $T$.

Tasks are connected by dependency relations encoded in a \emph{software network} $SN$.
\revtwo{Let $P_T\subset\tasks$ be a set of predecessor tasks for task $T\in\tasks$. If task $\hat T \in P_T$ (that is, $\hat T$ is a \emph{predecessor} of task $T$), } task $T$ can only be executed by a robot if the robot has data product $\data{\hat T}$.
If $\hat T$ is scheduled to be executed on the same robot as $T$, $\data{\hat T}$ is assumed to be available to $T$ as soon as the  computation of $\hat T$ is concluded.
If $\hat T$ and $T$ are scheduled on different robots, $\data{\hat T}$ must be transmitted from the robot executing $\hat T$ to the robot executing $T$ before execution of $T$ can commence.
An example of $SN$ used in our experiments is shown in Figure~\ref{fig:scenario_task_network}.

To ensure a solution exists, we require two assumptions.

\begin{assumption}[Feasibility]\label{assumption:isolated_feasibility}
There exists a schedule where all required tasks are scheduled.
\end{assumption}

\begin{assumption}[No circular dependencies]
The software network $SN$ does not have cycles.
\end{assumption}

\paragraph{Agents}
Agents in the network represent computing units.
Let there be $N\in\mathbb{Z}^{+}$ agents in the network. 
The agents are denoted by
$\agent{1},\thinspace \agent{2},\thinspace\ldots,\thinspace \agent{N}$.
Each agent has known on-board processing and storage capabilities.


The time and energy cost required to perform a task $T$ on agent $A_i$ are assumed to be known and denoted respectively as $\ctime{i}{T}$ and $\cenergy{i}{T}$.
Depending on the application, time and energy cost can capture the worst-case, expected, or bounded computation
time and energy; they are all considered to be deterministic. 


\begin{figure}[t]
\begin{centering}
\includegraphics[width=.6\columnwidth]{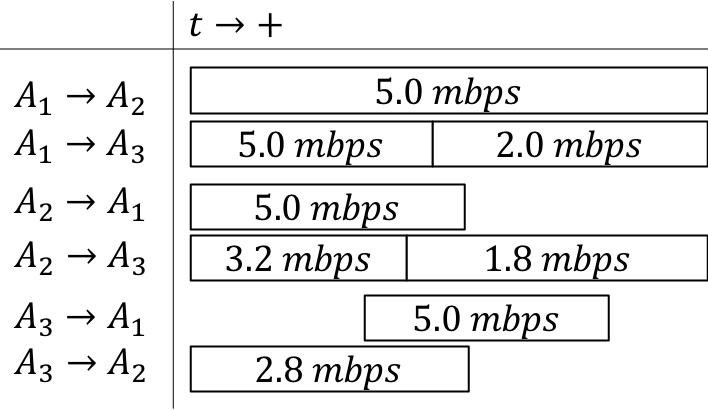}
\par\end{centering}
\caption{
  Contact graph for $3$ agents showing connectivity time windows and bandwidths available. \label{fig:Contact-graph}
}
\end{figure}

\paragraph{Contact Graph} \label{sec:cg}
Agents can communicate according to a prescribed time-varying \emph{contact graph} $CG$ which denotes the availability and bandwidth of communication links between the robots.

$CG$ is a graph with time-varying edges. 
Nodes $\mathcal{V}$ in $CG$ correspond to agents. 
For each time instant $k$, directed edges $\mathcal{E}_{k}$ model the availability of communication links; that is, $(i, j) \in \mathcal{E}_{k}$ if node $i$ can communicate to node $j$ at time $k$.
Each edge has a (time-varying) data rate ranging from $0$ (not connected) to $\infty$
(communicating to self), denoted by $\rate{i}{j}{k}$ for the rate from
\agent{i} to \agent{j} at time $k$.
An example timeline representation for $3$
agents with available bandwidths can be seen in
Figure~\ref{fig:Contact-graph}.

A key feature of DTN-based networking is Contact Graph Routing (CGR) \cite{wyatt2017,Araniti2015cg}.
CGR takes into account predictable link schedules and
bandwidth limits to automate data delivery and optimize the use of network
resources.
Accordingly, by incorporating DTN's store-forward mechanism
into the scheduling problem, it is possible to use mobile agents as
\emph{robotic routers} to ferry data packets between agents that are not directly connected.

Communicating the data product \data{T} from \agent{i} to \agent{j}
at time $k$ requires time
\rev{
\[
\ctime{ij}{T} = \min_{\tau} \left(\tau \text{ such that }  \int_{\kappa=k}^{k+\tau} \rate{i}{j}{\kappa} d \kappa \geq \size{T} \right),
\]
}


\rev{that is, $\ctime{ij}{T}$ is the shortest time required to transmit a total of $\size{T}$ bits at an instantaneous data rate $\rate{i}{j}{\cdot}$ starting at time $k$.}
\rev{If the data rate $\rate{i}{j}{\cdot}$ is constant through the communication window and sufficiently long for the transmission to occur,
 the expression can be simplified to
 $\ctime{ij}{T} = \size{T}/\rate{i}{j}{k}$.}

\revtwo{ We model agents as single-threaded computers. 
If a robot actually has multiple processors, even of different types, these can be accommodated by modeling each processor as a computing agent, and connecting physically co-located processors with infinite-bandwidth, zero latency communication links. }

\begin{assumption}[Computational resource availability]
\label{assumption:single_task}
Agents can only perform a single task at any given time, including transmitting or receiving data products.
\end{assumption}

\begin{assumption}[Communication self-loops]
Agents take $0$ time to communicate the solution to
themselves. 
\end{assumption}

\paragraph{Schedule}
A schedule is (a) a mapping of tasks to agents and start-times, denoted as
$\sol:T\rightarrow (\agent{i}, k)$ where $i\in[1,\ldots,N]$ and $k\geq0$, 
and (b) a list of inter-agent communications 
$(\agent{i}, \agent{j}, \data{T}, k)$ denoting the transmission of $\data{T}$ from $\agent{i}$ to $\agent{j}$ from time $k$ to time \rev{$k+\ctime{ij}{T} : \left(\int_{k}^{k+\ctime{ij}{T}} \rate{i}{j}{\kappa} d \kappa = \size{T}\right)$.}

\paragraph{Optimization Objectives}
We consider several optimization objectives (formalized in the following section), including:
\begin{itemize}
\item \emph{Optional tasks}: maximize the sum of the rewards $r(T)$ for optional tasks $T$ that are included in the schedule;
\item \emph{Makespan}: minimize the maximum completion time of all scheduled tasks; 
\item \emph{Energy cost}: minimize the sum of the energy costs $\cenergy{i}{T}$ for tasks included in the schedule;
\end{itemize}

\paragraph{Scheduling Problem}
We are now in a position to state the communication-aware  computation task scheduling problem for heterogeneous multi-robot systems.

\begin{problem}[Communication-Aware Computation Task Scheduling Problem for Heterogeneous Multi-Robot Systems]
  \label{prob:schedule}
  Given a set of tasks modeled as a software network $SN$, a list of
  computational agents \agent{i}, $i\in[1\ldots N]$, a contact graph $CG$,
  and a maximum schedule length $C^\star$, find a schedule that satisfies:
  \begin{enumerate}
  \item The maximum overall computation time is no more than $C^\star$;
  \item All required tasks $T\in \required$ are scheduled;
  \item A task $T$ is only scheduled on  agent $\agent{i}$ at time $k$ if the agent has received all the data product $\data{\hat T}$ for predecessor tasks $\hat T \in P_T$;
  \item Every agent performs at most one task (including transmitting and receiving data products) at any time;
  \item The selected optimization objective is maximized.
  \end{enumerate}
  \label{prob:sched}
\end{problem}

\paragraph{Notes on Problem Assumptions}

The assumption that a feasible schedule including all required tasks exists (Assumption \ref{assumption:isolated_feasibility}) is appropriate for multi-robot systems where each required task ``belongs'' to a specific robot (i.e., the task is performed with inputs collected by the robot, and the output of the task is to be consumed by the same robot). Examples of such tasks include localization, mapping, and path planning. In such a setting, it is reasonable to assume that each robot should be able to perform all of its own required tasks with no assistance from other computation nodes; on the other hand, cooperation between robots can decrease the makespan, reduce energy use, and enable the completion of optional tasks.

The contact graph is assumed to be known in advance. This assumption is reasonable  in many space applications, specifically in surface-to-orbit communications, orbit-to-orbit communications, and surface-to-surface communication in unobstructed environments, where the capacity of the communication channel can be predicted to a high degree of accuracy.  In obstructed environments where communication models are highly uncertain (e.g., subsurface voids such as caves, mines, tunnels)  a conservative estimate of the channel capacity could be used. Extending Problem \ref{prob:sched} to explicitly capture uncertainty in the communication graph is an interesting direction for future research. 

Finally, Problem \ref{prob:sched} also assumes that the communication graph is not part of the optimization process. The problem of optimizing the contact graph by prescribing the agents' motion is beyond the scope of this paper \revtwo{(for a good example of the vast literature see \cite{yan2013co,ghaffarkhah2011communication})}; note the tools described in this paper can be used as an optimization subroutine to numerically assess the effect of proposed changes in the contact graph on the performance of the multi-robot system.


\section{Scheduling Algorithm}
\label{sec:scheduler}
\subsection{ILP formulation}

\newcommand{\horizon}{\ensuremath{C^\star_d}}

\newcommand{\sn}{\ensuremath{SN}}

We formulate Problem~\ref{prob:sched} as an integer linear program (ILP). 
We consider a discrete-time approximation of the problem with a time horizon of $\horizon$ time steps, each of duration $C^\star/\horizon$, corresponding to the maximum schedule length $C^\star$. 
\revtwo{As is common in ILP formulations, the number of time steps can be set to any value that balances runtime vs granularity.}
The optimization variables are:
\begin{itemize}
\item $X$, a set of Boolean variables of size $N\cdot M \cdot \horizon$. $X(i,T,\dtime)$ is true if and only if agent $A_i$ starts computing task $T$ at time $\dtime$.
\item $D$, a set of Boolean variables of size $N \cdot M \cdot \horizon$. $D(i,T,\dtime)$ is true if and only if agent $A_i$ has stored the data products $d(T)$ of task $T$ at time $\dtime$. 
\item $C$, a set of Boolean variables of size $N^2 \cdot M \cdot \horizon$. $C(i,j,T,\dtime)$ is true if and only if agent $A_i$ communicates part or all of data products $\data{T}$ to agent $A_j$ at time $\dtime$.
\end{itemize}

The optimization objective $R$ can be expressed as follows:
\begin{itemize}
\item Maximize the sum of the rewards for completed optional tasks:
\begin{subequations}
\begin{equation}
R_r =  \sum_{i=1}^N \sum_{T\in\tasks\setminus \required}  \sum_{\dtime=1}^{\horizon-\ctime{i}{T}} r(T) X(i,T,\dtime) 
\end{equation}
\item Minimize the makespan of the problem:
\begin{equation}
R_M = -\max_{i\in[1, N]} \max_{T\in \tasks} \max_{\dtime\in[1, \horizon]} \left( \dtime + \ctime{i}{T} \right)X(i,T,\dtime)
\end{equation}
\item Minimize the energy cost of the problem:
\begin{equation}
R_e = - \sum_{i=1}^N \sum_{T\in\tasks}  \sum_{\dtime=1}^{\horizon} \cenergy{i}{T} X(i,T,\dtime) 
\end{equation}
\label{eq:MILP:costs}
\end{subequations}
\end{itemize}

We are now in a position to formally state the ILP formulation of Problem \ref{prob:schedule}:

{
\ifrelaxedv \else
\small
\fi
\begingroup
\allowdisplaybreaks
\begin{subequations}\label{eq:MILP}
\begin{flalign}
& \underset{X, D, C}{\text{maximize }} R \\
& \text{subject to}\nonumber \\
&  \sum_{i=1}^N   \sum_{\dtime=1}^{\horizon-\ctime{i}{T}} X(i,T,\dtime)  = 1 \quad \forall T\in \required \label{eq:MILP:requiredtasks}\\
&  \sum_{i=1}^N \sum_{\dtime=1}^{\horizon-\ctime{i}{T}} X(i,T,\dtime)  \leq 1  \quad \forall T\in\tasks\setminus \required \label{eq:MILP:optionaltasks}\\
&  X(i,T,\dtime) \leq D(i,L,\dtime) \label{eq:MILP:prereqs} \\
&\quad  \forall i\in[1,\ndots,N], T\in[1,\ndots,M], L\in P_T, \dtime\in[1,\ndots,\horizon] \nonumber \\
& \sum_{T=1}^M \left[ \sum_{j=1}^N\left( C(i,j,T,\dtime) + C(j,i,T,\dtime) \right) + \sum_{\mathclap{\hat \dtime = \max(1,\dtime-\ctime{i}{T})}}^\dtime X(i,T,\hat \dtime) \right] \nonumber \\
& \quad \leq 1  \quad \quad \forall i\in[1,\ndots,N], \dtime\in[1,\ndots,\horizon] \label{eq:MILP:computation}\\
& D(i,T,\dtime+1)\!-\!D(i,T,\dtime)  \nonumber \\
& \quad \leq \sum_{\tau=1}^\dtime \sum_{j=1}^N \frac{r_{ji}(\tau)}{\size{T}} C(j,i,T,\tau) + \sum_{\tau=1}^{\mathclap{\dtime-\ctime{i}{T}}} X(i,T,\tau) \nonumber \\
& \quad  \forall i\in[1,\ndots,N],  T\in[1,\ndots,M], \dtime\in[1,\ndots,\horizon-1] \label{eq:MILP:learning}\\
& C(i,j,T,\dtime) \leq D(i,T,\dtime) \nonumber \\
& \quad  \forall  i,j \in[1,\ndots,N], T\in[1,\ndots,M], \dtime\in[1,\ndots,T] \label{eq:MILP:knowledge}\\
& D(i,T,1)=0 \quad \forall i \in[1,\ndots,N], T\in[1,\ndots,M] \label{eq:MILP:initialinfo}
\end{flalign}
\end{subequations}
\endgroup
}
Equation \eqref{eq:MILP:requiredtasks} ensures that all required tasks are
performed and \eqref{eq:MILP:optionaltasks} that optional tasks are performed
at most once. 

Equation \eqref{eq:MILP:prereqs} requires that agents only
start a task if they have access to the data products of all its predecessor
tasks. 
Equation \eqref{eq:MILP:computation} captures the agents' limited computation
resources by enforcing Assumption \ref{assumption:single_task}. 
Equation \eqref{eq:MILP:learning} ensures that agents learn the content of a
task's data products only if they (i) receive such information from other
agents (possibly over multiple time steps, each carrying a fraction $r_{ij}(\dtime)/\size{T}$ of the data product) or (ii) complete the task themselves.
Equation \eqref{eq:MILP:knowledge} ensures that agents only communicate a data
product if they have stored the data product themselves. Finally, Equation
\eqref{eq:MILP:initialinfo} models the fact that data products are initially
unknown to all agents.

The ILP has $N^2M\horizon+2NM\horizon$ Boolean variables and $M(N(3\horizon-1)+N )+N\horizon$
constraints; instances with dozens of agents and tasks and horizons of 50--100 time steps can be readily solved by
state-of-the-art ILP solvers, as shown in Section \ref{sec:experiment}. 

\revtwo{
\subsection{Modeling Extension: Capturing Network Interference}
}
The ILP formulation can be extended to capture network interference as follows.
In \eqref{eq:MILP}, link bandwidths $r_{ij}$ are assumed to be fixed and independent of each other: that is, the communication bandwidth $r_{ij}$ on a link is assumed to be achievable regardless of communication activity on other links. This may not hold for systems with robots in close proximity that share the same wireless channel.
In such a setting, interference introduces a coupling between the achievable bandwidths on different links, and the overall amount of data that can be exchanged by interfering links is limited by the \emph{channel capacity} of the shared physical medium.

The formulation in \eqref{eq:MILP} can be extended to capture a first-order approximation of this effect, letting individual link bit rates be decision variables subject to constraints on the overall channel capacity.
Effectively, agents are allowed to use less than the full capacity of individual links to ensure that their transmissions do not cause interference on other links sharing the same wireless channel.

To accommodate, define an additional set of real-valued decision variables $R$ of size $N^2 \cdot M \cdot \horizon$. $R(i,j,T,\dtime)$ denotes the amount of bits of the data product of task $T$ that is transmitted from agent $A_i$ to agent $A_j$ in time interval $\dtime$.

Under this model, 
the interfering links' channel capacity $r(\interferenceset, \dtime)$  (that is, the overall amount of bits that links in $\interferenceset$ can simultaneously transmit) is known, for each discrete time interval $\dtime$ and each subset $\interferenceset \in \interferencesets \subset 2^{N^2}$ of links that is subject to mutual interference.
In order to avoid introducing an exponential number of constraints, it is desirable to consider a modest number of sets of interfering links \revtwo{based on the agents' geographical proximity}. For instance, if all robots are operating in close proximity and can interfere with each other, the overall bandwidth of \emph{all} links should be constrained to be smaller than the capacity of the shared channel, \revtwo{resulting in the addition of a single interference constraint}.

Equation \eqref{eq:MILP:learning} is replaced by the following equations:
{
\ifrelaxedv \else
\small
\fi
\begin{subequations}
\begin{flalign}
& R(i,j,T,\dtime) \leq r_{i,j}(\dtime) C(i,j,T,\dtime) \nonumber \\ 
&\quad \forall  i,j \in[1,\ndots,N], T\in[1,\ndots,M], \dtime\in[1,\ndots,T] \label{eq:MILP:boolean_bandwidth} \\
& D(i,T,\dtime+1)\!-\!D(i,T,\dtime)  \nonumber \\
& \quad \leq \sum_{\tau=1}^\dtime \sum_{j=1}^N \frac{1}{\size{T}} R(j,i,T,\tau) + \sum_{\tau=1}^{\mathclap{\dtime-\ctime{i}{T}}} X(i,T,\tau) \nonumber \\
& \quad  \forall i\in[1,\ndots,N],  T\in[1,\ndots,M], \dtime\in[1,\ndots,\horizon-1] \label{eq:MILP:learning_interference} \\
& \sum_{i,j \in {i}} \sum_{T\in[1,\ndots,M]} \!\!\!R(j,i,T,\dtime) \leq r(i,\dtime) \quad \forall i\in I, \dtime\in[1,\ndots,T] \label{eq:MILP:channel_capacity}
\end{flalign}
\end{subequations}
}

Equation \eqref{eq:MILP:boolean_bandwidth} ensures that the effective bit rate on a link is nonzero only if a communication occurs on the link; Equation \eqref{eq:MILP:learning_interference} models the process by which robots learn data products through communication, closely following Equation \eqref{eq:MILP:learning}; and Equation \eqref{eq:MILP:channel_capacity} ensures that the sum of all effective bit rates on interfering links does not exceed the channel capacity.



\subsection{Distributed, real-time implementation \label{sec:ilp:distributed}}
\begin{figure}[h]
\centering
{
\jvspace{-1em}
\includegraphics[width=\columnwidth]{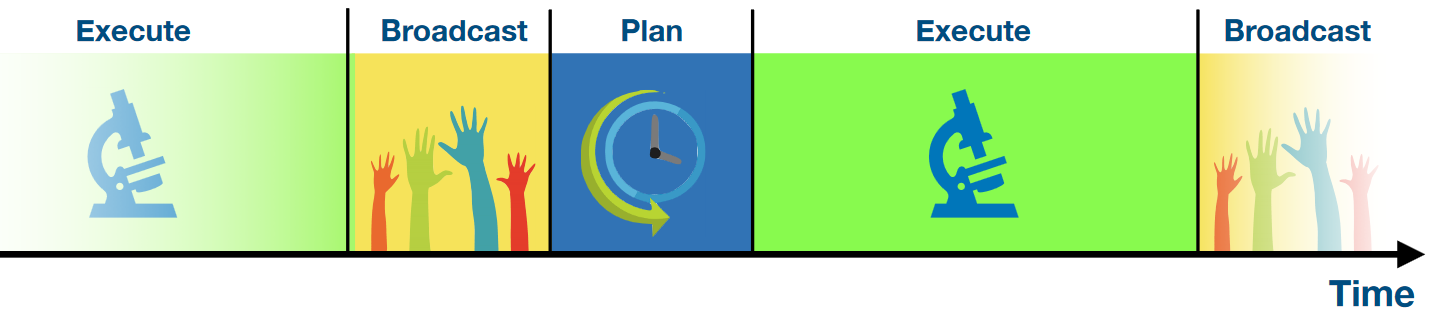}
}
\jvspace{-1em}
\caption{Distributed implementation of the ILP relies on a broadcast-plan-execute cycle. First, agents exchange information about their own state through
a message-passing algorithm and achieve a consensus on the system state. Next, all agents solve Problem \eqref{eq:MILP} with the system state
as input and with a deterministic stopping criterion. Finally, all agents execute the tasks assigned to them by the solution to Problem \eqref{eq:MILP}.} 
\label{fig:broadcast-plan-execute}
\jvspace{-.5em}
\end{figure}

In order to provide a \emph{distributed}, \emph{real-time} implementation of the scheduler
presented above suitable for field use, we leverage a shared-world approach using a ``broadcast, plan, and execute'' cycle (shown in Figure 
\ref{fig:broadcast-plan-execute}). 

Agents are assumed to have access to a common clock and have pre-existing knowledge of the duration of the broadcast, plan, and execute phases of the cycle.
The agents also know what programs or processes may be included in the software network \sn (even if not all agents can execute all processes). They are not aware of the optimization objective, namely, the execution times, energy costs, sequences, and rewards of individual tasks. 

\paragraph{Broadcast} At an agreed-upon time, agents start the ``broadcast'' phase; during this phase, agents 
exchange their state
with all other agents through a flooding message-passing algorithm \cite[Ch. 4]{lynch1996distributed}, and achieve a consensus on the overall system state.
The duration of the broadcast phase is selected to ensure that consensus can be achieved for any possible network topology. As discussed in \ifextendedv
Appendix \ref{apx:flooding_clustering}
\else
the Extended Version \cite{Hook2021EV}
\fi
, if the communication network is strongly connected, systems with 10-50 agents can achieve consensus in under a second under conservative assumptions on the size of agents' states and available link bandwidths.

The state of each agent includes
(i)  the estimated present and future bandwidths $r_{ij}$ between each agent and their neighbors,
(ii) the time and energy costs $\{\ctime{i}{T}\}_{i\in [1, N], T \in \tasks}$, $\{\cenergy{i}{T}\}_{i\in [1, N], T \in \tasks}$ required by the agent to perform each possible task, and
(iii) the rewards $\{r(T)\}_{T\in\tasks\setminus \required}$ for performing optional tasks.

This approach is responsive to time-varying task rewards and agent capabilities. 
However, the choice of a single broadcast epoch per round
does cause some delay in responsiveness, since agent capabilities and rewards can only be updated
if they appear prior to the start of the broadcast phase for each cycle.

\paragraph{Plan}
Once the broadcast phase is over, agents switch to the ``plan'' phase.
In this phase, each agent solves Problem~\eqref{eq:MILP} 
with the network topology, tasks set, and vehicle states computed in the broadcast phase as inputs.

Problem~\ref{prob:sched} 
is in general NP-hard, and a solver may fail to find an optimal solution within the allocated time.
To ensure that a feasible final solution is found, we provide the solver with a trivial initial solution (which exists, according to Assumption \ref{assumption:isolated_feasibility}). 
To ensure that all agents agree on the same solution, we use a \emph{deterministic} MILP solver (i.e., a solver that explores the decision tree according to a deterministic policy), and we employ a deterministic stopping criterion (i.e., the solver terminates after a prescribed, deterministic number of branch-and-bound steps, selected to ensure termination within the duration of the ``plan'' phase).
  
\paragraph{Execute} Once the plan phase is over, agents switch to the execution phase; here, each agent reads the output of Problem \eqref{eq:MILP} and executes the tasks that
are assigned to itself according to the timing prescribed by the schedule.

This approach provides a \emph{distributed} and \emph{anytime} implementation of 
Problem~\ref{prob:sched} which we implement and test in the next Section.

\subsection{Remarks}

\rev{
In the problem formulation, communication tasks do consume computational resources on both the transmitter and the receiver (Assumption \ref{assumption:single_task}).
This is in line with the current mode of operations of space missions, where communication is not concurrent with other activities due to computational, power, and reliability considerations. As a result, the rovers' activities should be \emph{synchronized}: in absence of synchronization, a rover's transmission could interrupt computational activities on the receiver, or be lost if the receiver is unavailable.

In the light of this, the selection of a "broadcast-plan-execute" distributed implementation, which relies on the availability of synchronization between the agents, is preferable for its simplicity and ease of verification.

In cases where agents can concurrently communicate and perform computational tasks, a more versatile \emph{asynchronous} distributed implementation could also be used. We propose such an asynchronous load-sharing execution mechanism in \cite{RossiVaqueroEtAl2020}; the proposed architecture is agnostic to the task-allocation mechanism used, and is therefore compatible with plans provided by the ILP.
}

The broadcast-plan-execute cycle relies on synchronization of the agents' clocks and on accurate knowledge of the duration of tasks to be executed. Deviations from predicted execution times can result both in tasks not being completed in the ``execute'' phase, and in missed communication windows (if, e.g., a task is not completed by the time its data products should be transmitted to another agent). The cyclic nature of the broadcast-plan-execute cycle allows ``missed'' tasks to be rescheduled at a later time step; nevertheless, the design of \emph{robust} scheduling algorithms that can accommodate uncertainty in synchronization and in task execution times is an interesting direction for future research.

While the flooding-based synchronization mechanism itself is quite robust (as discussed in the previous subsection),  the overall scheduling approach is \emph{not} robust to failures of the broadcasting synchronization mechanism. 
The integration of more robust coordination mechanisms (e.g., challenge-response to verify that agents have achieved a consensus, and  watchdogs triggering
the execution of an agreed-upon contingency plan) are interesting directions for future research.

Finally, the complexity of the ILP scales exponentially with the number of agents;
accordingly, in principle, it may be infeasible to obtain a high-quality solution to Problem \ref{eq:MILP} at a sufficient cadence for control of a multi-robot system, especially on embedded platforms.
However, in Section~\ref{sec:experiment:benchmark}, we show that state-of-the-art ILP solvers can provide high-quality (if not optimal) solutions within tens of seconds, even on highly constrained platforms. 


\begin{figure*}[htb]
  \begin{center}
    \includegraphics[height=4.5cm]{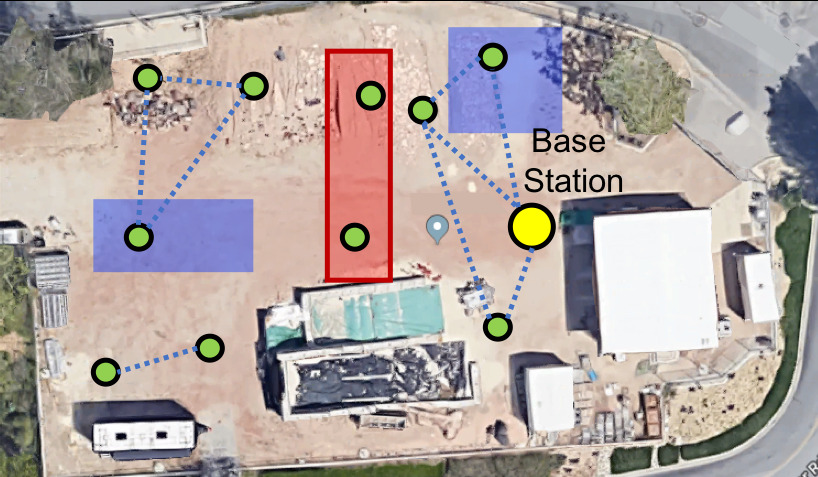}
    \includegraphics[height=4.5cm]{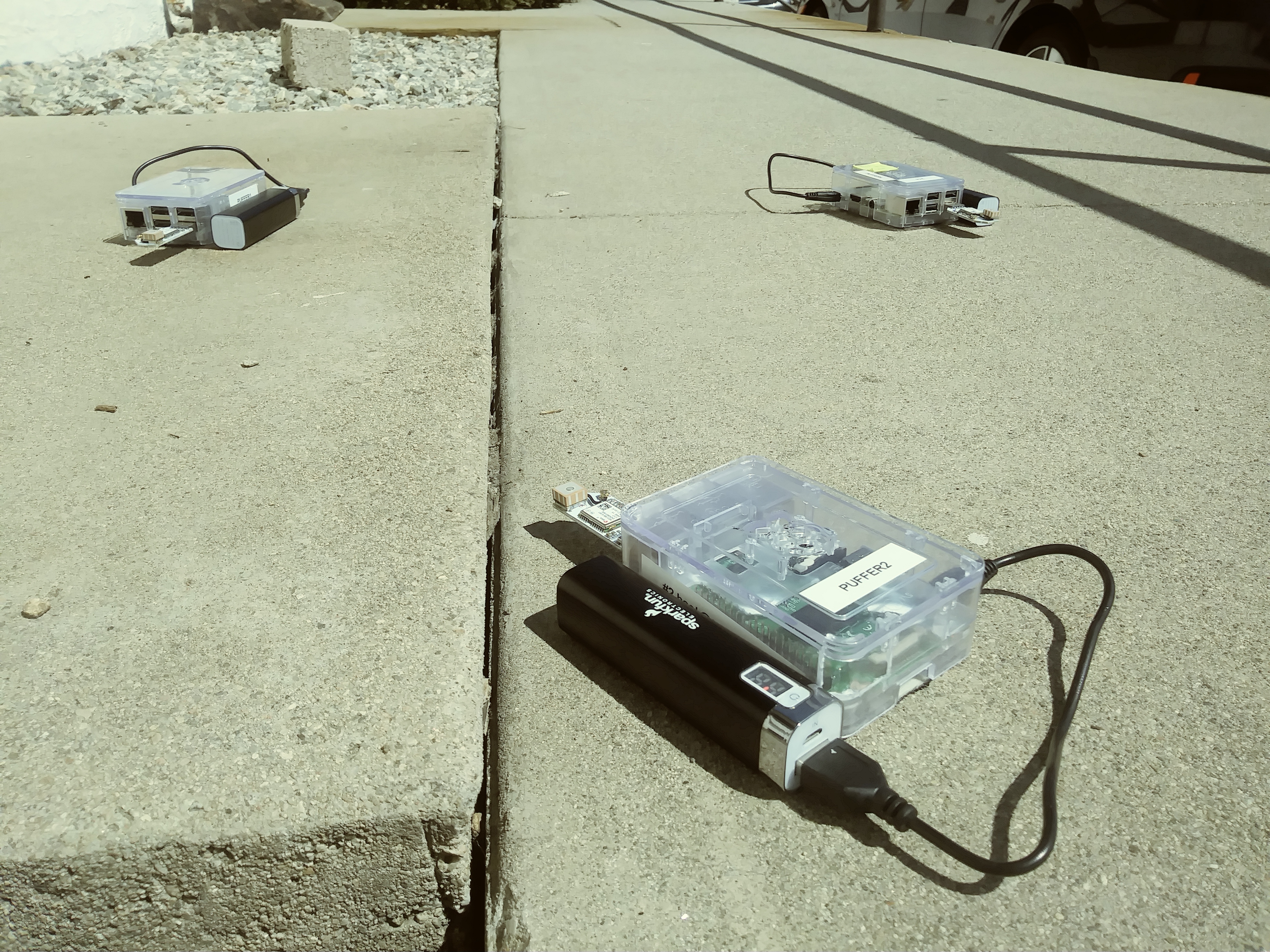}
    
    \vspace{.5cm}
    
    \includegraphics[height=4.4cm]{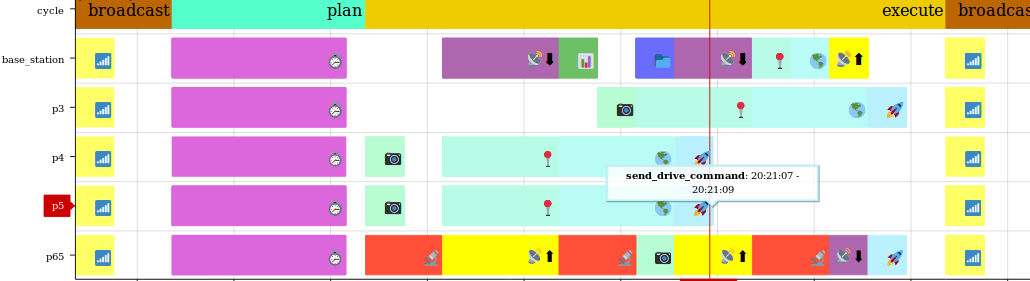}
    
  \end{center}
  \caption{Illustrative scenario in the Mars Yard at JPL (top left), pictures of the hardware nodes (top right), and one scheduled timeline (bottom).
   Timelines represents
 the operational cycle and the task allocation. Communication links can be disabled
 to test system adaptation and relocation of tasks. RVIZ view provides vehicle
 positioning and network topology information.}
    \label{fig:mars_yard_scenario}
 \label{fig:viz_tools}
 \jvspace{-1.25em}
\end{figure*}


\section{Experiments} \label{sec:experiment}



\begin{figure}[thb]
\centering
\includegraphics[width=.8\columnwidth]{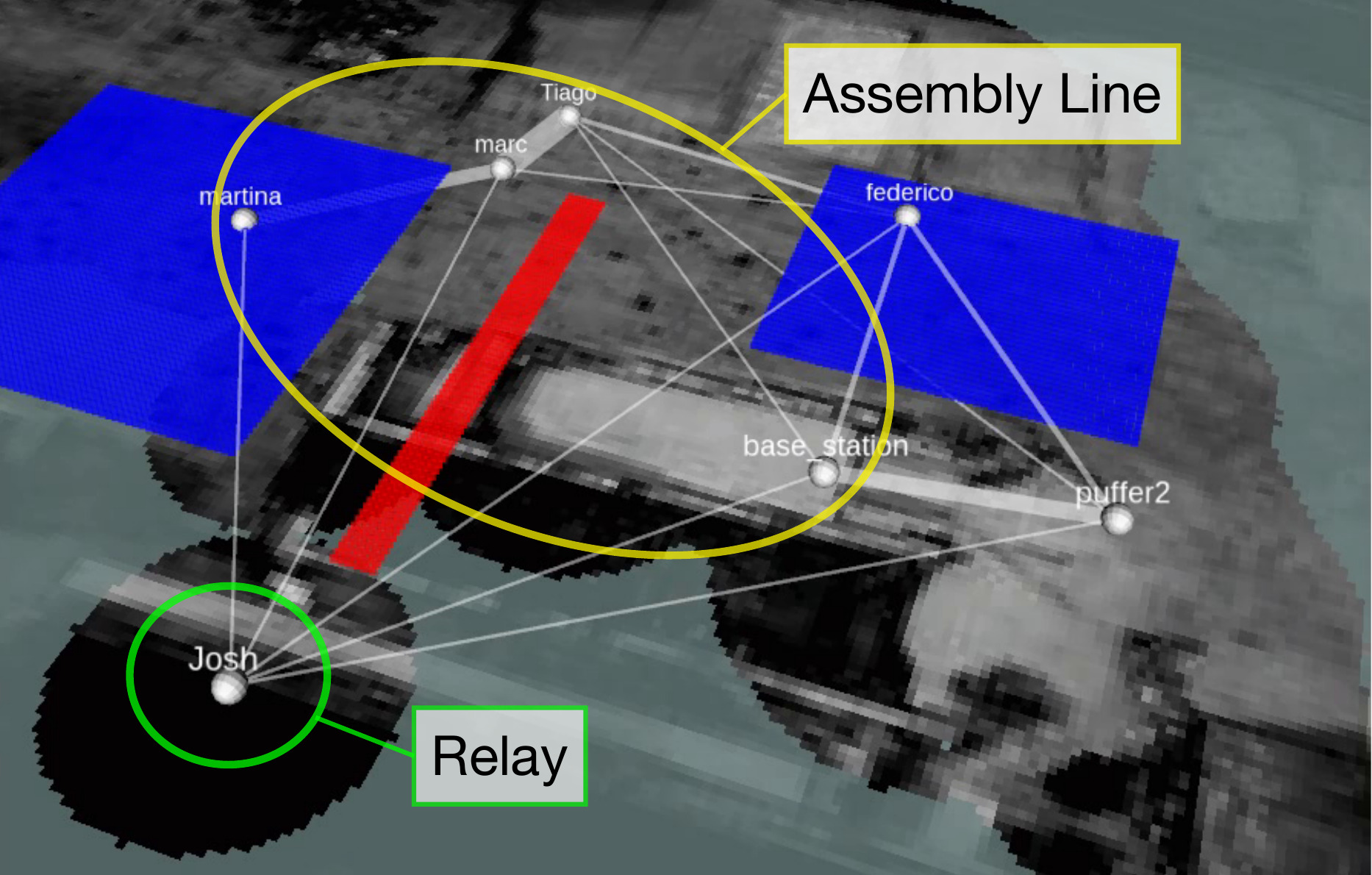}
\caption{Relay and assembly line emerging behaviors (Yellow and green annotations were added manually to the RViz output from the field demonstration).}
\label{fig:relay_assembly_line_behavior}
\end{figure}

\begin{figure*}[thb]
\centering
\includegraphics[width=.75\textwidth]{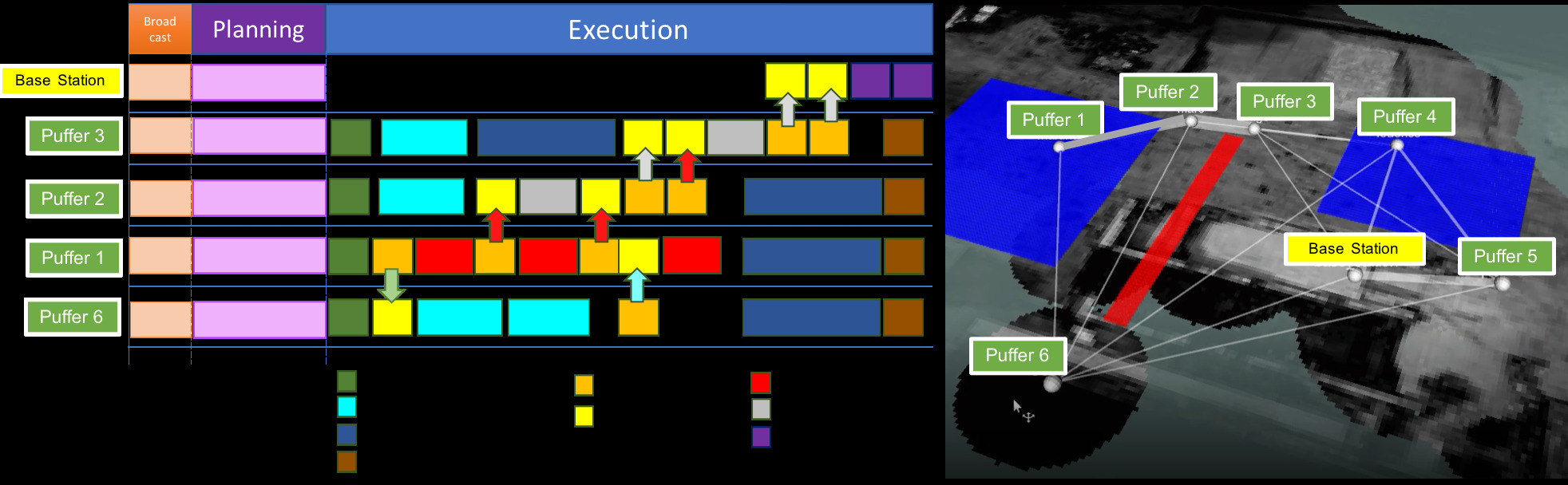}
\caption{Illustrative example of the assembly line case.}
\label{fig:assembly_line_scenario}
\end{figure*}

\begin{figure*}[htb]
\centering
  \includegraphics[width=\textwidth,trim={2cm 0 0.25cm 13cm},clip]{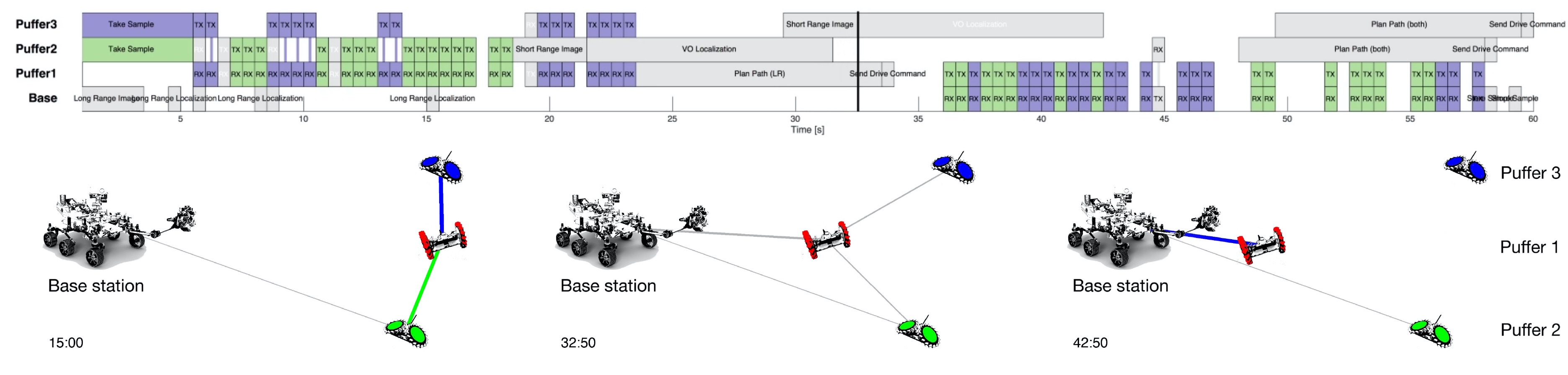}
  \caption{Example simulation of the data mule scenario. Left: Three Puffers have a weak link to the base station, but the middle robot will move closer and so both robots transmit their data to it rather than directly to the base station. Right: Later, the red robot transmits all data to the base station. See supplementary videos and Section~\ref{sec:reproducing_results}.}
\label{fig:data_mule_scenario}
\jvspace{-1.25em}
\end{figure*}

\rev{In this section, we explore the performance of the proposed approach on a
variety of realistic problems.  First, we present field tests of a
\emph{distributed} implementation on a set of mobile, wirelessly-connected,
Raspberry Pis.  Second, we assess the computational complexity and performance
of the approach through rigorous benchmarks on a variety of computational
architectures.}

\subsection{Distributed hybrid implementation}

\rev{The goals of these experiments were to implement and test the
\emph{distributed} version of Problem \eqref{eq:MILP}, and, specifically, the
broadcast-plan-execute architecture in Section \ref{sec:ilp:distributed}, in
realistic and challenging environments.  We sought to check for four important
characteristics of a field-deployed system. 

\begin{itemize}
  \item \revtwo{\emph{Robustness}: Does the proposed approach run for extended periods of time? It may crash, we may
    encounter violated assumptions, or the ILP may not find a feasible solution
    in time.}
  \item \emph{Computational cost}: Does the implementation scale well and run quickly on
    realistic computing architectures?
  \item \emph{Networking}: Are communication tasks scheduled reasonably, despite the
    additional complexity of scheduling computation? Since our ILP contains
    data routing as a sub-problem, we expect the solutions to contain reasonable
    routing behaviors.
  \item \emph{Load Balancing}: Does the solution exhibit load balancing behaviors when nodes
    with uneven computational load have good communication between them?
  \item \emph{Science Optimization}: Does the system achieve an increase in
    throughput of science data compared to a naive approach? 
\end{itemize}}

We used a notional multi-robot scenario where multiple small
rovers perform both ``housekeeping'' tasks (e.g., sensing, path planning) and
science tasks (e.g., microscope measurements) and are aided by a
computationally capable base station. 
This is illustrated in Figure \ref{fig:mars_yard_scenario}.
\rev{The concept of operations is loosely based on JPL's PUFFER robots 
\cite{karras2017pop}}.

\rev{The software network used is shown in Figure~\ref{fig:scenario_task_network}}.
Tasks are arranged in two sets. ``Housekeeping'' tasks (Figure
\ref{fig:scenario_task_network}, top)
are based  on the Mars Perseverance rover's
autonomy architecture, and their execution time is based on actual benchmarks on 
Perseverance's on-board RAD750 \cite{riebertalk}.
Housekeeping tasks include (i) capturing an image of the terrain,
(ii) self-localization based on that image, (iii) planning a path through the
environment, and (iv) dispatching the drive command.  While image capture and
drive command have to be executed on board, localization
and path planning tasks can be delegated to another robot in the network.

To model optional, autonomous science activities, we also added the ``science tasks'' shown in
the bottom of Figure~\ref{fig:scenario_task_network}.  Specifically, PUFFERs 
can (i) collect a sample from the environment, (ii) analyze it,  and (iii) send the
analysis data to the base station for storage and eventual uplink.
The sample analysis task can be assigned to another node.
Only agents inside pre-designated ``science zones'' can perform sample collection;
 storage must be performed by the base station.
Each science task has a reward associated to it; the reward for sample collection is set to 5, the reward for
data analysis is  10, and the reward for storing data is 20.
\rev{Note that no actual sampling and analysis tasks were executed; rather,
task execution was simulated by allocating time in the schedule computed by each node.}

This set of ``science tasks'' represents the scenario where PUFFERs explore a distributed
but spatially-correlated phenomenon, such as water moisture levels, by
performing kriging \cite{kriging}, a process routinely used for spatial
estimation in farming on Earth~\cite{tokekar2016sampletspn}.

The base station's computational power
is an order of magnitude larger than an individual robot's, and it is equipped with the same
communication equipment as the other nodes in the network.
The base station is not assigned any required tasks; its key role is to serve as a supporting node for
sharing the computational load of the network.

\rev{In the field experiments, the PUFFERs were represented by Raspberry Pis (model 3) with a GPS receiver, and
the base station was a desktop computer at a fixed location.  The Pis were
moved about within an outdoors experimental area with
two marked ``science zones'' by human experimenters.
We had limited control over positions of the nodes during the
experiment and demonstration, due to the use of the highly
portable Raspberry Pis and participation of enthusiastic observers from JPL and
direction from observing sponsors.  
Accordingly, this experiment was an ideal test of the \emph{reliability} and \emph{robustness} of the overall architecture;
separate software benchmarks (reported in the next section) are better suited to assess
the computational cost and performance of the approach.}

To control the communications network, all platforms were connected through a
WiFi router; bandwidths between nodes were computed in simulation based on (i)
inter-agent distance, (ii) the presence of no-communication zones (shown in red
in Figure \ref{fig:viz_tools}), and (iii) direct human intervention
\rev{(i.e., the experimenters' ability to disable selected communication links on demand)}.  Data
rates between agents were based on inter-agent range according to a piecewise
constant function: available bandwidth ranged from 1 Mbps for ranges between
15-200 meters to 11 Mbps for ranges between 0-5 meters, or zero (unavailable)
if the line-of-sight between two vehicles crossed obstructions (shown as a red
zones in Figure \ref{fig:viz_tools}).

The logical layers on all agents was implemented using the Robot Operating
System (ROS). For each vehicle in the network, we developed a set of simulated
components, including a communication system to handle the broadcasting phase
and transfer of data, a science component to simulate science tasks, a
navigation/localization system to keep track of the vehicle's pose, and a
controller to plan and dispatch tasks according to the scheduler.

We also developed tools to  visualize 1) the execution timeline (and task
allocation), 2) the network topology and bandwidth, and 3) the vehicles'
positions in the environment. The tools also enable manipulation of the
simulation by adding additional no-communication zones and removing
communication links.  The software interface is shown in
Figure~\ref{fig:viz_tools}.

The agents were assumed to have access to a common clock, and we considered an
operational cycle 45 seconds long, where 5 seconds was dedicated to the
``broadcast'' phase, 10 seconds the ``plan'' phase, and 30 seconds to execution
(in line with the expected duration of the tasks in the software network in
Figure~\ref{fig:scenario_task_network}). \revtwo{In practice, we would expect a much longer execution cycle, and perhaps shorter broadcast / plan phase, but these values allowed increased introspection and iteration during field tests. }

The ILOG CPLEX solver was used to solve the ILP. Since CPLEX does not support
an ARM architecture, we deployed the scheduler in an x86 server (AWS m5.xlarge)
that was queried at each plan phase by each vehicle.  Vehicles called the
scheduler independently; the solution received by each vehicle was guaranteed
to be consistent with the other vehicles' through use of a deterministic solver
with a deterministic stopping criterion (as discussed in
Section~\ref{sec:ilp:distributed}).  The  deterministic amount of solver steps
corresponded to approximately 10 seconds of execution on the Pi. To ensure
anytime availability of a feasible solution, the solver was seeded with an
initial solution where agents did not share any computational tasks and
executed no optional tasks, which is guaranteed to be feasible. 

\rev{Representative portions of the field test are shown in a video in the
Supplementary Material\footnote{Available at
\url{https://youtu.be/zTQ7Y4-ax2A}}.  
We provide a software release for full exploration of the
results (Section ~\ref{sec:reproducing_results}).}
Experiments were run for four hours, and demonstrated all of the following
characteristics. 

\rev{\paragraph{Robustness}

During the 4-hour long demonstration, nodes were added and removed from
the network (by activating and deactivating the corresponding Raspberry Pi's),
 and active nodes were moved around by observers,
including in and out of science zones. We verified that the proposed
approach is able to \emph{consistently} provide good solutions to problems
with 3 to 15 nodes within the 10-seconds planning window, and that the
broadcast-plan-execute architecture can be used to provide a distributed
implementation of Problem \eqref{eq:MILP} that is robust to unforeseen,
human-driven changes in the network topology and in the tasks to be scheduled.

\paragraph{Networking and Data Relay}

One of the most intuitive, and obviously beneficial emergent behaviors we observed was that
of relay activities.  Relay nodes, informally speaking, did nothing more than 
 relay communications between other nodes while tending to their own housekeeping tasks. 
This behavior was induced reliably through the use of the no-communication zone to
block direct communication with the base station. Traffic was reliably routed though nodes that were
between the base station and the sender instead, as shown in Figure~\ref{fig:relay_assembly_line_behavior}. 

\paragraph{Load Balancing and Science Clusters}

The choice of the software network places additional load on nodes that are in
``science zones'', by adding (optional) science tasks to their list of tasks.
In a recurring behavior, ``science clusters'' formed whenever one vehicle was inside
a science region, and other vehicles were nearby but outside.
For example in Figure \ref{fig:relay_assembly_line_behavior}, 
The nodes in both science zones off-loaded its localization and path planning
tasks to the other nearby agent, so as to perform multiple sample collection and
analysis tasks. 

\paragraph{Science Optimization}

Due to the timing chosen for the software network, the proposed approach could yield at
most a threefold increase in the number of optional science tasks performed for each
node in a science zone. That is, the sum of the computation times of all
relocatable housekeeping tasks was twice the cost of a science task: therefore, by doing
\emph{only} science and offloading all relocatable housekeeping tasks, an agent could gather
three times more science than would have been possible with no load sharing.
The additional analysis and storage tasks placed additional load on nodes
\emph{outside} of the science zone, if appropriately tasked.  This threshold was achieved for some nodes that had sufficient nearby nodes, and sufficient throughput to the base station. 
Again in Figure \ref{fig:relay_assembly_line_behavior}, the left science node
was able to schedule three sample-gather tasks, by offloading tasks to nearby
agents.  We can explore the likelihood of this occurring in random networks in Section~\ref{sec:experiment:benchmark}.

\paragraph{Science Optimization with Assembly Lines}

The combination of relay and load balancing produces and interesting result
that was unintended but obvious in hindsight.  When the system did reach the
maximum observed science throughput, the relay nodes also served as
computational aids for the science tasks, analyzing the data enroute to the
base station akin to an ``assembly line''.  We illustrate an occurrence of such
a case in Figure \ref{fig:assembly_line_scenario}.} The node labelled PUFFER 1
is in the left-most science zone and offloads localization (cyan) to nearby
PUFFER 6, as in the ``science cluster'' scenario.  PUFFER 1 also schedules
three samples (red).  Two sample data products are then transferred to PUFFER
2, which acts as a relay to the base station.  PUFFER 2 analyzes one sample and
forwards the resulting analysis result and one sample data product to PUFFER 3;
PUFFER 3 analyzes the sample and transfers two analyzed data products to the
base station for storage.  The third sample data product is not analyzed or
stored due to the short time horizon; nevertheless, it is collected to receive
the corresponding reward. As mentioned, a threefold increase per node in
science zones is the maximum possible increase due to the amount of time to
execute compared to the time costs of all the tasks. 

\rev{
 
The ``assembly line'' result is quite interesting and may have unexplored
efficiency increases for edge computing networks like terrestrial 5G networks. 

\paragraph{Store and Forward, and Data Muling} 
Because the planner has knowledge of the future state of the communications
network, it should be possible to plan for future connectivity and
store-and-forward packets to a node in preparation for a link coming online. If
the link comes online because the storing node \emph{moves}, this is sometimes
called ``data muling'' \cite{bhadauria2011robotic}.  
We did not observe this in field testing because we could not predict the
future state of the communications network, due to human manipulation.
However, the data muling behavior was readily observed and reproduced in 
simulation, as shown in Figure \ref{fig:data_mule_scenario} and in the video
in the Supplementary Material.}

\subsection{Software benchmarks}
\label{sec:experiment:benchmark}

\begin{figure}
\includegraphics[width=\columnwidth]{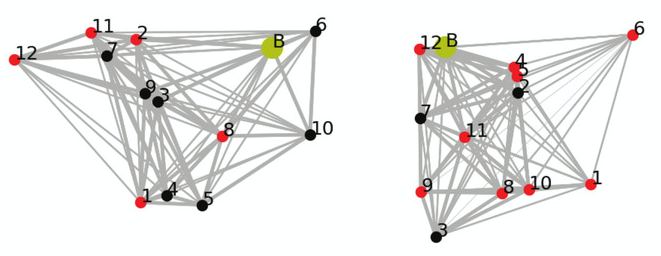}
\caption{Two example scenarios from the numerical experiments. The base station is yellow. Nodes able to perform science tasks are red; nodes unable to perform science tasks are shown in black, the width of edges shows bandwidth. }
\label{fig:experiments:instances}
\end{figure}

Next, we show through numerical
results that the proposed ILP can be solved efficiently on a variety of
hardware platforms, including embedded platforms suitable for robotics
applications, \rev{and we explore the benefits of the  approach compared to a ``selfish'' scenario where agents cannot share computational tasks.}
To this end, we test the performance of a \emph{centralized} version of Problem \eqref{eq:MILP} on several hardware architectures for twenty randomly-generated network topologies
(shown in Figure \ref{fig:experiments:instances})
and several cost functions. \rev{In each scenario, a subset of the agents was randomly placed in ``science zones''; 
agents in science zones were able to collect one sample, which could optionally be analyzed and stored.}

For each instance, the number of agents (proportional to number of tasks to schedule) was varied from 2 to 13 agents  to assess the scalability of the proposed approach. Optimization objectives included (i) maximization of reward from optional task, (ii) minimization of energy expenditure, and (iii) a linear combination of the two.
The problem was solved on several computing platforms, specifically:
\begin{itemize}
\item a modern Intel Xeon workstation equipped with a 10-core  E5-2687W processor;
\item an embedded Qualcomm Flight platform equipped with a APQ8096 SoC;
\item a Powerbook G3 computer equipped with a single-core PowerPC 750 clocked at 500 MHz, the same CPU (albeit without radiation tolerance adjustments) as the RAD750 used on the Curiosity and Mars 2020 rovers \cite{Bajracharya2008autonomy}.
\end{itemize}

\begin{figure*}[t]
\centering
\includegraphics[width=.3\textwidth]{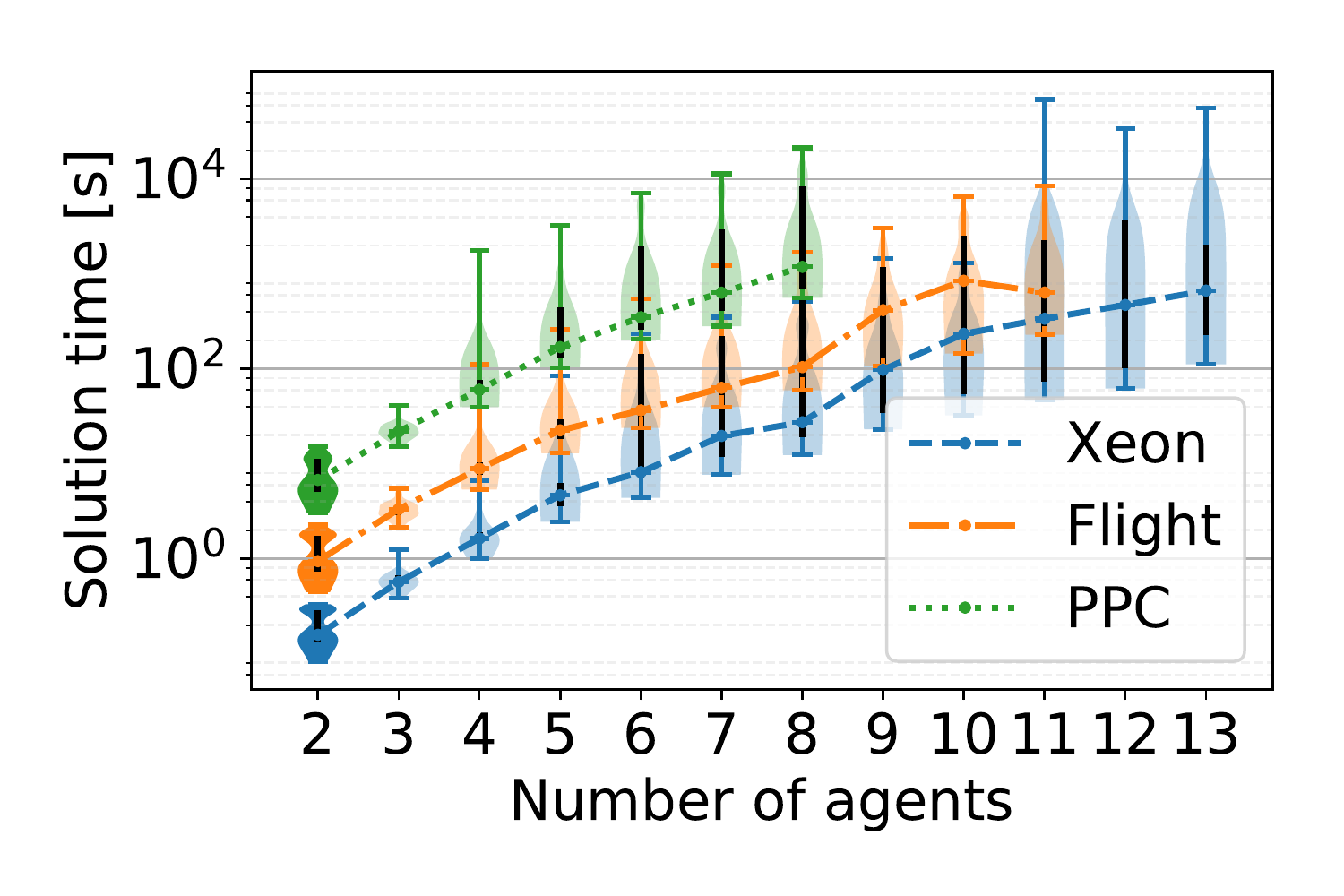}
\includegraphics[width=.3\textwidth]{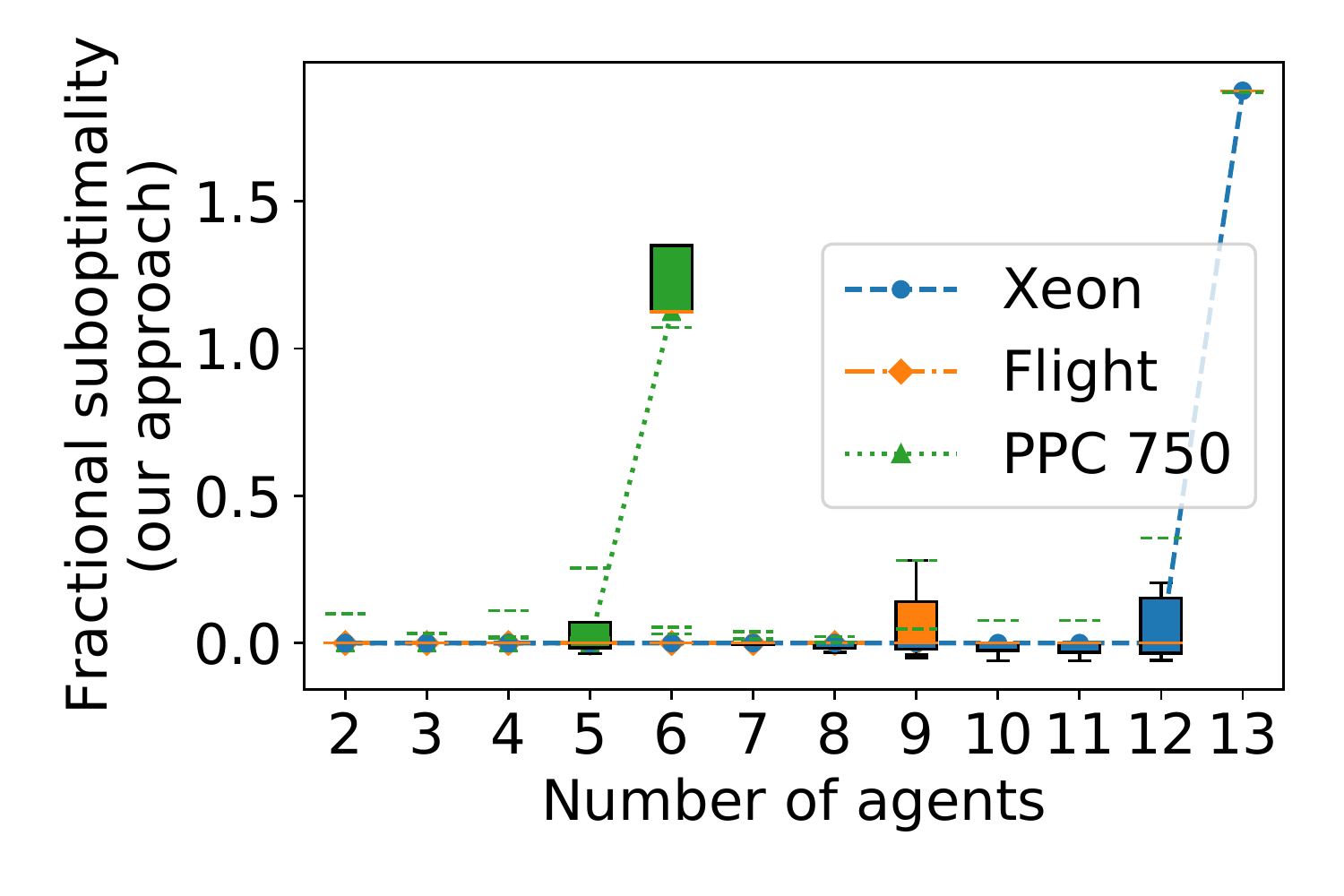}
\includegraphics[width=.3\textwidth]{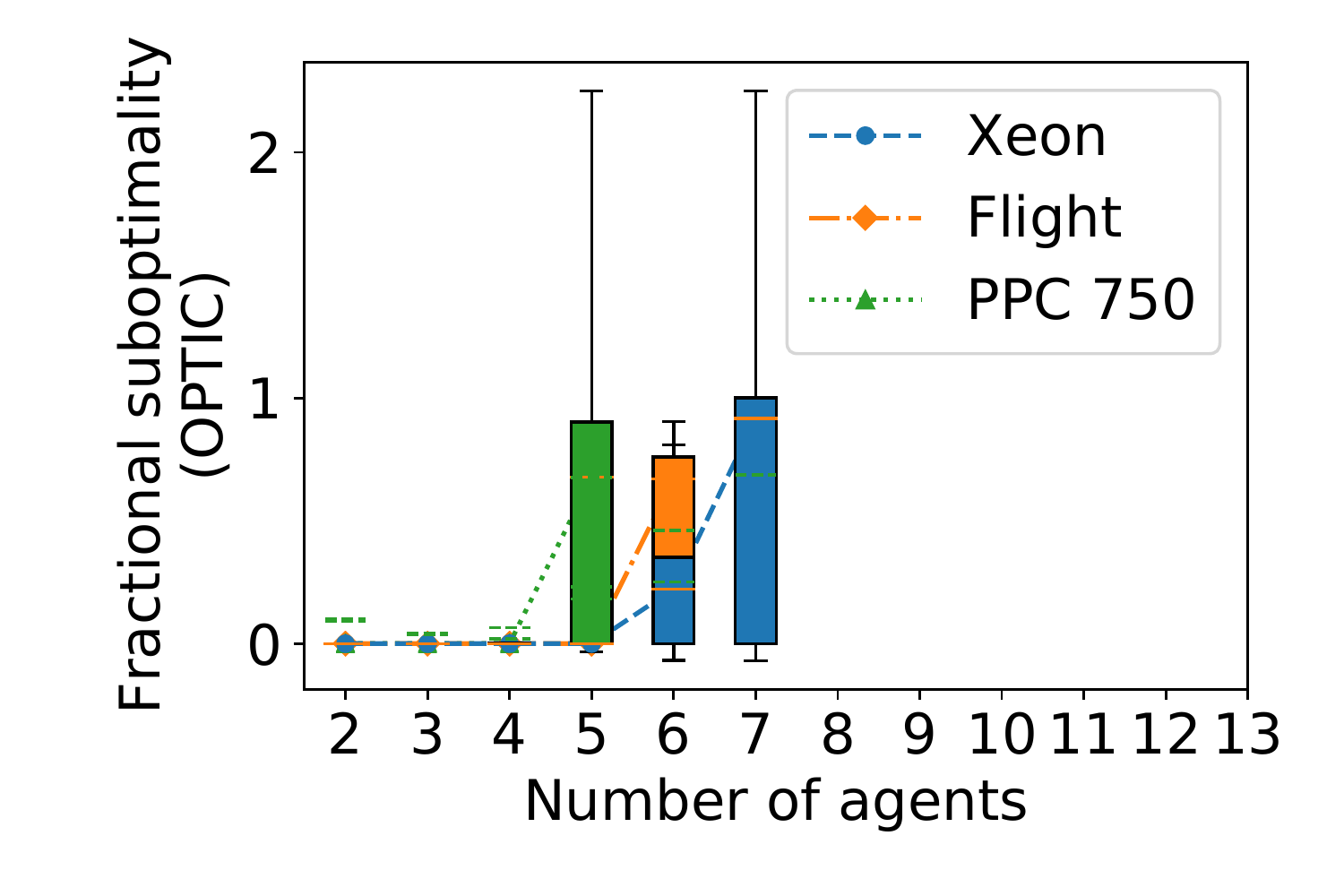}
\jvspace{-.5em}
\caption{Numerical results on several hardware platforms, and comparison with the OPTIC PDDL scheduler. Left: Time required to solve Problem \ref{eq:MILP} to optimality. Middle: Suboptimality as a fraction of the optimal solution after 60s of execution for Problem \eqref{eq:MILP} and Right: for the OPTIC PDDL solver \cite{benton2012temporal}.}
\label{fig:experiments:benchmark}
\jvspace{-1em}
\end{figure*}

The ILP was solved with the SCIP solver \cite{GleixnerEtal2018OO}. For each problem, we computed both the time required for the solver to find and certify an optimal solution, and the quality of the best solution obtained after 60 seconds of execution. We also compared the performance of the proposed scheduler with the state-of-the-art OPTIC PDDL scheduler \cite{benton2012temporal}. Results are shown in Figure \ref{fig:experiments:benchmark}.

On the Xeon architecture, the median solution time for problems with up to 6 agents is under 10s, and the median solution time for problems with up to 9 agents is under 100s (Figure \ref{fig:experiments:benchmark}, top). The proposed anytime implementation is consistently able to find an optimal solution for problems with up to 11 agents in under 60s (Figure \ref{fig:experiments:benchmark}, middle).
On the embedded Qualcomm SoC, the median solution time for problems with up to 4 agents is under 10s, and the anytime implementation finds the optimal solution to problems with up to 8 agents in under 60s. 
Finally, even the highly limited PPC 750 processor is able to find an optimal solution to problems with up to 5 agents in under 60s - a remarkable achievement for a 20-year-old processor.

The ILP scheduler offers superior performance compared to the anytime implementation of the OPTIC scheduler (Figure \ref{fig:experiments:benchmark}, bottom). In particular, solving Problem \eqref{eq:MILP} results in higher-quality solutions for a given problem size and execution time, and OPTIC is unable to return solutions for problems with more than 7 agents even on the Xeon architecture.

\rev{We also assessed the potential benefits of the proposed approach on a more complex version of the problem, where each agent in a ``science zone'' was able to collect up to \emph{three} samples. We solved the same set of problems shown in Figure \ref{fig:experiments:instances} with up to nine agents; for each instance, we compared the solution to the ILP with a ``selfish'' allocation where agents were not allowed to share computational tasks (except for the storage task, which was constrained to be executed on the base station).
We evaluated the solution quality both after 60s of execution, and after 3600s of execution (a time sufficient to achieve and prove optimality for the vast majority of the scenarios considered). 
\begin{figure}[t]
\centering
\includegraphics[width=\ifextendedv0.7\else0.48\fi\columnwidth]{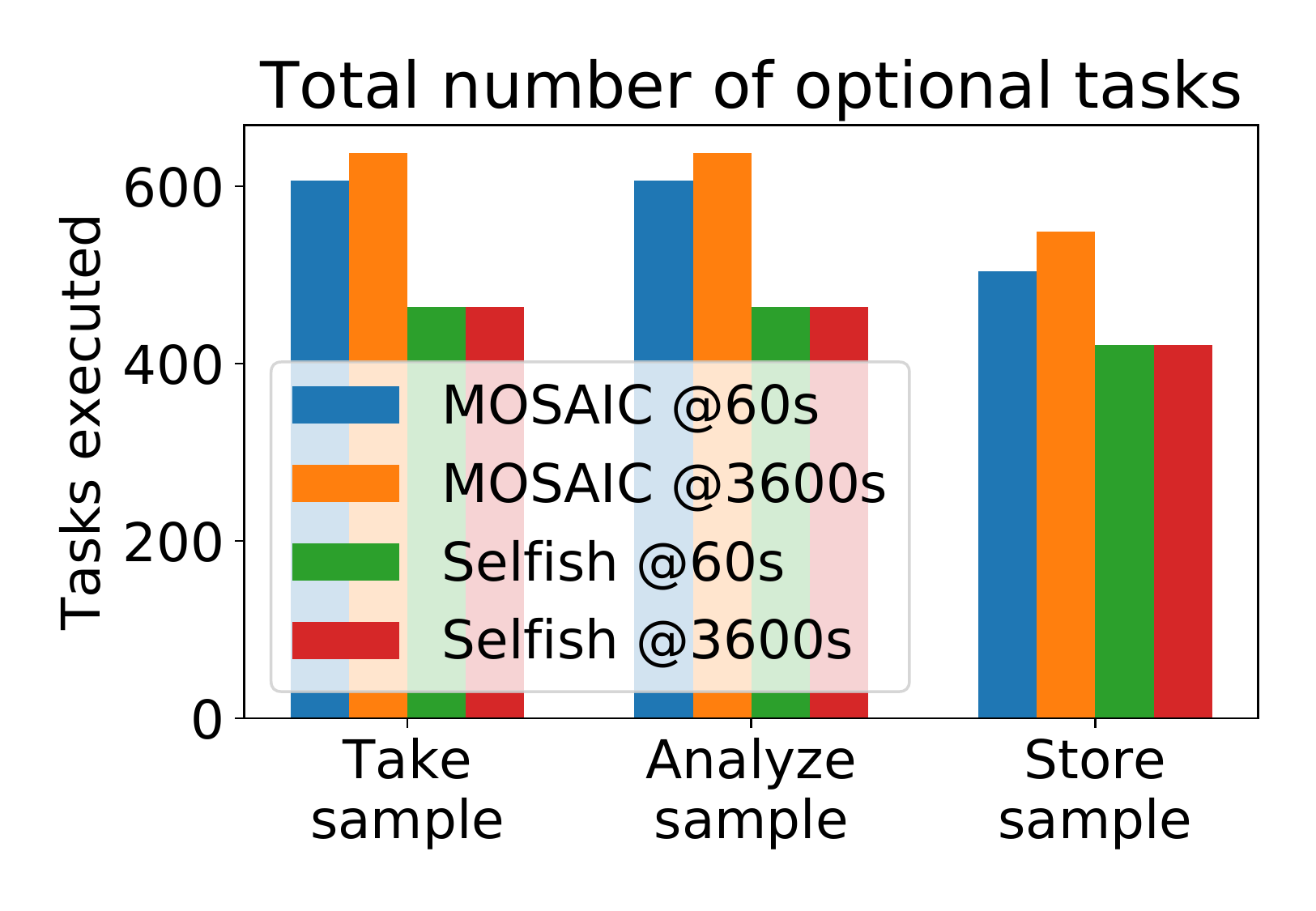}
\includegraphics[width=\ifextendedv0.7\else0.48\fi\columnwidth]{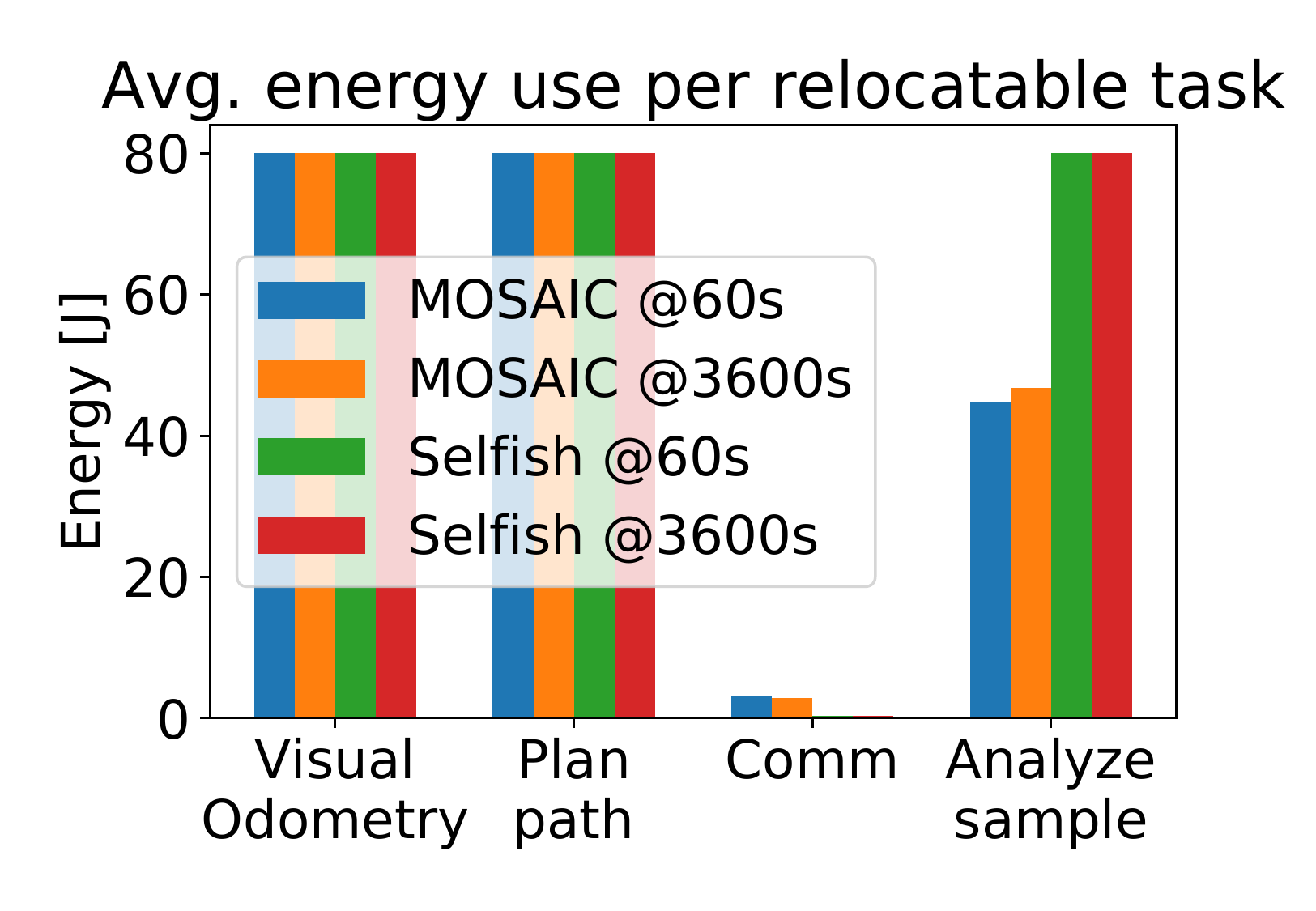}
  \caption{Proposed approach vs  ``selfish'' approach (no sharing of cpu time). Left: tasks performed. Right: average energy per task.}
\label{fig:experiments:performance}
\end{figure}

Figure \ref{fig:experiments:performance} shows the overall number of tasks performed and the average energy usage per task across all problem instances. After 1h of execution, the proposed approach yields a 37.3\% increase in the number of samples collected and analyzed, and a 30.4\% increase in the number of samples stored, compared to the selfish approach; the approach also results in a 41.4\% reduction in average energy use for the sample analysis task, which more than outweighs the small increase in energy use for communications. Remarkably, a similar trend is observed even when the solver is stopped after 60s: here, the proposed approach results in a 30.6\% increase in the number of samples collected and analyzed, a 19.7\% increase in the number of samples stored, and a 44\% decrease in the average energy use for sample analysis compared to the selfish approach.

Collectively, these results show that the proposed approach holds promise to yield significant increases in scientific returns and decreased energy usage; can be implemented on embedded robotic architectures with modest computational performance; and performs well in highly dynamic environments, making it well-suited for field robotics multi-agent applications.
}

\paragraph{Reproducing Our Results}
\label{sec:reproducing_results}

We have released implementations of Problem \eqref{eq:MILP} using
the CPLEX, SCIP, and GLPK MILP solvers under a permissive open-source license.
The implementations are available online at \url{github.com/nasa/mosaic}.
Provided scenario files allow reproduction of all of the
emergent behaviors discussed.


\section{Conclusion}\label{sec:disc}


In this paper, we described the communication-aware computation task scheduling problem for heterogeneous multi-robot system and the Multi-robot On-site Shared Analytics Information and Computing (MOSAIC) architecture. We proposed an ILP formulation that allows to optimally schedule computational tasks in heterogeneous multi-robot systems with time-varying communication links.
We showed that the ILP formulation is amenable to a distributed implementation; can be solved efficiently  on embedded computing architectures; and can result in a threefold increase in science returns compared to systems with no computational load-sharing.

A number of directions for future research are of interest.
First, we plan to explore pathways to infusion of the MOSAIC architecture in future multi-robot planetary exploration missions.
Proposed Mars Sample Return mission concepts plan to re-visit the same area with multiple launches to fetch, retrieve, and eventually launch soil samples for return to Earth \cite{mattingly2011msr}. 
This offers an especially attractive avenue for deployment of MOSAIC, where each deployed asset could act as an ``infrastructure upgrade'', providing communication, computation, and data analysis services for all subsequent assets. Agents participating in the MOSAIC could include Cubesats similar to MarCO
\ifextendedv
 \cite{hodges2016marco,schoolcraft2016marco};
 \else
 \cite{schoolcraft2016marco};
 \fi
  assets  embedded in the ``sky crane'' lander and dropped during the ``flyaway'' phase
\ifextendedv
\cite{korzun2010skycrane,sell2013powered};
\else
\cite{korzun2010skycrane};
\fi
tethered balloons \cite{kerzhanovich2004balloon}; and aerostationary orbiters providing constant assistance to half the Mars surface 
\ifextendedv
\cite{breidenthal2016design,breidenthal2018space}.
\else
\cite{breidenthal2018space}.
\fi
The algorithms proposed in this paper can be used during the system design phase to optimize the hardware of the
distributed missions by simulating the scheduling problem in the loop with an iterative hardware trade space explorer such as \cite{herzig2017tradespace}. 

Second, we will design software libraries and middlewares that enable integration
of the proposed scheduler with existing autonomy software, autonomously and 
transparently distributing computational tasks according to the optimal schedule.
A preliminary effort in this direction can be found in 
\cite{RossiVaqueroEtAl2020}.

Finally, it is of interest to extend the proposed scheduling approach to handle uncertainty in the contact graph and in the execution time of individual task.
One promising research avenue is to incorporate stochastic optimization tools as well as probabilistic planning and scheduling
approaches \cite{santana-vaquero-et-al-2016} to the computation sharing problem,
which hold promise to provide guarantees that the MOSAIC is able to operate within
given bounds on the uncertainty of the problem inputs.

\jvspace{-.5em}
\section*{Acknowledgements}
This work was carried out at the Jet Propulsion Laboratory, California Institute of Technology, under a contract with the National Aeronautics and Space Administration. \\


\jvspace{-1em}
{\small
\bibliographystyle{IEEEtran} 
\bibliography{RBT-VanderhookJ.31.bib}
}

\ifextendedv{
\jvspace{-1em}
\begin{appendix}

\subsection{Common Knowledge through Flooding and Clustering.}
\label{apx:flooding_clustering}

The distributed approach relies on a flooding algorithm to ensure that all agents achieve common knowledge of the system state (specifically, the network state, the agents' capabilities, and the optional task rewards).
In this Appendix we show that, if the communication network is strongly connected for the duration of the "broadcast" phase, the flooding algorithm can achieve consensus in well under 1 second for systems with a moderate number of agents (10-50) through an order-of-magnitude, worst-case analysis.

Consider a system with $N$ agents. The maximum number of relay hops that any message must traverse in order to reach any agent is $N-1$.
In a flooding algorithm, each agent relays each received message to every neighbor once; accordingly, at most $N$ messages are sent on each communication link. If the size of a message containing an agent's state is $b$ bits, and all communication links provide a bandwidth of $r$ (neglecting interference), the maximum time required to achieve consensus is $N(N-1) b/r$.

The size of an agent's state can be very compact.  

Link bandwidth is typically negotiated by the communication protocol among a limited number of options; for instance, the 802.11g WiFi specification allows up to eight possible bandwidths \cite{van1999new}, and state-of-the-art UWB radios offer up to three data rates \cite{Decawavedwm1000}.
Accordingly, we assume that the bandwidth of a link can be represented by 8 discrete levels, or 3 bits; the bandwidths of all neighbors of an agent can be encoded in at most $3N$ bits.

In typical applications, an agent's computational capabilities are chiefly determined by its battery state-of-charge, which can be discretized in a small number of buckets for scheduling purposes. We assume each agent's capabilities can be represented through 8 discrete levels, or 3 bits.

Finally, rewards for optional tasks typically assume a small set of possible values (e.g., 1 if the task can be performed, and 0 if it is not).
We assume that each agent can choose to schedule 10 possible tasks and task rewards can be represented by four possible levels or 2 bits, for a total of 20 bits. 

Accordingly, the overall size of an agent's state is $b=3N+23$ bits, and the time required to achieve consensus is upper-bounded by $N(N-1)(3N+23)/r$. 
For a system of 10 agents, even modest 5kbps radio links are sufficient to ensure that the flooding algorithms terminates within 0.95s; a system of 50 agents equipped with 1Mbps links (a reasonable rate for close-range operations) is able to achieve consensus within 0.4s

For systems with large number of agents or systems with very low communication bandwidth where flooding is prohibitive, we advocate the use of a \emph{distributed clustering} approach. In such an approach, agents use either a label-propagation algorithm which finds clusters of well-connected nodes \cite{RaghavanAlbertSoundar2007} or a modified version of the GHS distributed MST algorithm \cite{RossiPavone2013} which creates a forest of well-connected trees of bounded height to self-organize into clusters. Within each cluster, agents solve Problem \ref{eq:MILP} through the broadcast-plan-execute cycle.  By organizing in relatively small, well-connected clusters, the agents can both ensure termination of the flooding algorithm within the ``broadcast'' phase, and reduce the computational complexity of Problem \ref{eq:MILP}, at the price of a (typically small) loss of optimality. 
The exploration of such an approach is an interesting direction for future research.

\end{appendix}
\fi

\end{document}